\documentclass[10pt, a4paper]{article}
\usepackage{lrec}
\usepackage{multibib}
\newcites{languageresource}{Language Resources}
\usepackage[utf8]{inputenc}
\usepackage[T1]{fontenc}
\usepackage{graphicx}
\usepackage{tabularx}
\usepackage{soul}
\usepackage{booktabs}
\usepackage{multicol}

\usepackage{epstopdf}

\usepackage{hyperref}
\usepackage{xstring}

\usepackage{xspace}

\def\aleda{{\sf Aleda}\xspace}

\newcommand{\citet}[1]{\citeauthor{#1} \shortcite{#1}}
\newcommand{\citep}{\cite}

\newcommand{\camembert}{CamemBERT\xspace}

\newcolumntype{P}[1]{>{\RaggedRight\hspace{0pt}}p{#1}}

\newcommand{\ra}[1]{\renewcommand{\arraystretch}{#1}}

\title{Establishing a New State-of-the-Art for French Named Entity Recognition}

\name{Pedro Javier Ortiz Su\'arez$^{1,2}$, Yoann Dupont$^{1,2}$, Benjamin Muller$^{1,2}$,\\
\large\textbf{Laurent Romary$^1$, Beno\^it Sagot$^1$}}

\address{Inria$^1$, Sorbonne Universit\'e$^2$.\\
         2 rue Simone Iff, 75012 Paris$^1$, 21 rue de l’École de médecine, 75006 Paris$^2$.\\
         \{pedro.ortiz, benjamin.muller, laurent.romary, benoit.sagot\}@inria.fr\\
         yoa.dupont@gmail.com\\}

\abstract{
The French TreeBank developed at the University Paris 7 is the main source of morphosyntactic
and syntactic annotations for French. However, it does not include explicit information related
to named entities, which are among the most useful information for several natural language
processing tasks and applications. Moreover, no large-scale French corpus with named entity
annotations contain referential information, which complement the type and the span of each
mention with an indication of the entity it refers to. We have manually annotated the French
TreeBank with such information, after an automatic pre-annotation step. We sketch the underlying
annotation guidelines and we provide a few figures about the resulting annotations.\\ \newline \Keywords{Named Entity Recognition, French, Language Modeling}
}

\begin{document}

\maketitleabstract

\section{Introduction}
\label{sec:intro}

Named entity recognition (NER) is the widely studied task consisting in identifying text spans that denote {\em named entities} such as person, location and organization names, to name the most important types. Such text spans are called named entity {\em mentions}. In NER, mentions are generally not only identified, but also classified according to a more or less fine-grained ontology, thereby allowing for instance to distinguish between the telecommunication company {\em Orange} and the town {\em Orange} in southern France (amongst others). Importantly, it has long been recognised that the type of named entities can be defined in two ways, which underlies the notion of metonymy: the intrinsic type ({\em France} is always a location) and the contextual type (in {\em la France a signé un traité} `France signed a treaty', {\em France} denotes an organization).

NER has been an important task in natural language processing for quite some time. It was already the focus of the MUC conferences and associated shared tasks
\cite{marsh1998muc}, and later that of the CoNLL~2003 and ACE shared tasks \cite{conll03,doddington2004automatic}. Traditionally, as for instance was the case for the MUC shared tasks, only person names, location names, organization names, and sometimes ``other proper names'' are considered. However, the notion of named entity mention is sometimes extended to cover any text span that does not follow the general grammar of the language at hand, but a type- and often culture-specific grammar, thereby including entities ranging from product and brand names to dates and from URLs to monetary amounts and other types of numbers.

As for many other tasks, NER was first addressed using rule-based approaches, followed by statistical and now neural machine learning techniques (see Section~\ref{subsec:sota} for a brief discussion on NER approaches). Of course, evaluating NER systems as well as training machine-learning-based NER systems, statistical or neural, require named-entity-annotated corpora.
Unfortunately, most named entity annotated French corpora are oral transcripts, and they are not always freely available. The ESTER and ESTER2 corpora (60 plus 150 hours of NER-annotated broadcast transcripts)
\cite{ester_inter09}, as well as the Quaero
\cite{grouin2011law} corpus are based on oral transcripts (radio broadcasts). Interestingly, the Quaero corpus relies on an original, very rich and structured  definition of the notion of named entity \cite{rosset11}. It contains both the intrinsic and the contextual types of each mention, whereas the ESTER and ESTER2 corpora only provide the contextual type.

\newcite{sagot:hal-00703108} describe the addition to the French Treebank (FTB) \citelanguageresource{ftbLR} in its FTB-UC version \citelanguageresource{ftbucLR} of a new, freely available annotation layer providing named entity information in terms of span and type (NER) as well as reference (NE linking), using the Wikipedia-based \aleda \cite{sagot12aleda} as a reference entity database. This was the first freely available French corpus annotated with referential named entity information and the first freely available such corpus for the written journalistic genre. However, this annotation is provided in the form of an XML-annotated text with sentence boundaries but no tokenization. This corpus will be referred to as FTB-NE in the rest of the article.

Since the publication of that named entity FTB annotation layer, the field has evolved in many ways. Firstly, most treebanks are now available as part of the {\em Universal Dependencies} (UD)\footnote{\url{https://universaldependencies.org}} treebank collection. Secondly, neural approaches have considerably improved the state of the art in natural language processing in general and in NER in particular. In this regard, the emergence of contextual language models has played a major role. However, surprisingly few neural French NER systems have been published.\footnote{We are only aware of the {\em entity-fishing} NER (and NE linking) system developed by Patrice Lopez, a \href{https://github.com/kermitt2/entity-fishing}{freely available} yet unpublished system.} This might be because large contextual language models for French have only been made available very recently \cite{arxiv19camembert}. But it is also the result of the fact that getting access to the FTB with its named entity layer as well as using this corpus were not straightforward tasks. 

For a number of technical reasons, re-aligning the XML-format named entity FTB annotation layer created by \newcite{sagot:hal-00703108} with the ``official'' version of the FTB or, later, with the version of the FTB provided in the Universal Dependency (UD) framework was not a straightforward task.\footnote{Note that the UD version of the FTB is freely downloadable, but does not include the original tokens or lemmas. Only people with access to the original FTB can restore this information, as required by the  intellectual property status of the source text.} Moreover, due to the intellectual property status of the source text in the FTB, the named entity annotations could only be provided to people having signed the FTB license, which prevented them from being made freely downloadable online.

The goal of this paper is to establish a new state of the art for French NER by (i)~providing a new, easy-to-use UD-aligned version of the named entity annotation layer in the FTB and (ii)~using this corpus as a training and evaluation dataset for carrying out NER experiments using state-of-the-art architectures, thereby improving over the previous state of the art in French NER. In particular, by using both FastText embeddings \cite{bojanowski-etal-2017-enriching} and one of the versions of the CamemBERT French neural contextual language model \cite{arxiv19camembert} within an LSTM-CRF architecture, we can reach an F1-score of 90.25, a 6.5-point improvement over the previously state-of-the-art system SEM \cite{dupont2017exploration}.

\section{A named entity annotation layer for the UD version of the French TreeBank}

In this section, we describe the process whereby we re-aligned the named entity FTB annotations by \newcite{sagot:hal-00703108} with the UD version of the FTB. This makes it possible to share these annotations in the form of a set of additional columns that can easily be pasted to the UD FTB file. This new version of the named entity FTB layer is much more readily usable than the original XML version, and will serve as a basis for our experiments in the next sections.
Yet information about the named entity annotation guidelines, process and results can only be found in \newcite{sagot:hal-00703108}, which is written in French. We therefore begin with a brief summary of this publication before describing the alignment process.

\subsection{The original named entity FTB layer}
\label{subsec:originalannotations}

\newcite{sagot:hal-00703108} annotated the FTB with the span, absolute type\footnote{
  Every mention of {\em France} is annotated as a {\tt Location} with subtype {\tt Country}, as given in \aleda database, even if in context the mentioned entity is a political organization, the French people, a sports team, etc.}, sometimes subtype and \aleda unique identifier of each named entity mention.\footnote{Only proper nouns are considered as named entity mentions, thereby excluding other types of referential expressions.} Annotations are restricted to person, location, organization and company names, as well as a few product names.\footnote{More precisely, we used a tagset of 7 base NE types: {\tt Person}, {\tt Location}, {\tt Organization}, {\tt Company}, {\tt Product}, {\tt POI} (Point of Interest) and {\tt FictionChar}.} There are no nested entities. Non capitalized entity mentions (e.g.~{\em banque mondiale} `World Bank') are annotated only if they can be disambiguated independently of their context. Entity mentions that require the context to be disambiguated (e.g.~{\em Banque centrale}) are only annotated if they are capitalized.
  \footnote{So for instance, in {\em université de Nantes} `Nantes university', only {\em Nantes} is annotated, as a city, as {\em université} is written in lowercase letters. However, {\em Université de Nantes} `Nantes University' is wholly annotated as an organization. It is non-ambiguous because {\em Université} is capitalized. {\em Université de Montpellier} `Montpellier University' being ambiguous when the text of the FTB was written and when the named entity annotations were produced, only {\em Montpellier} is annotated, as a city.}
  For person names, grammatical or contextual words around the mention are not included in the mention (e.g.~in {\em M.~Jacques Chirac} or {\em le Président Jacques Chirac}, only {\em Jacques Chirac} is included in the mention).

Tags used for the annotation have the following information:
\begin{itemize}
\item the identifier of the NE in the \aleda database ({\tt eid} attribute); when a named entity is not present in the database, the identifier is {\tt null},\footnote{Specific conventions for entities that have merged, changed name, ceased to exist as such (e.g.~{\em Tchequoslovaquie}) or evolved in other ways are described in \newcite{sagot:hal-00703108}.}
\item the normalized named of the named entity as given in \aleda; for locations it is their name as given in GeoNames and for the others, it is the title of the corresponding French Wikipedia article,
\item the type and, when relevant, the subtype of the entity.
\end{itemize}
Here are two annotation examples:\\
\noindent{\small\tt <ENAMEX type="Organization" eid="1000000000016778"
name="Confédération française démocratique du travail">CFDT</ENAMEX>\\
<ENAMEX type="Location"
sub\_type="Country"
eid="2000000001861060" name="Japan">Japon</ENAMEX>}

\newcite{sagot:hal-00703108} annotated the 2007 version of the FTB treebank (with the exception of sentences that did not receive any functional annotation), i.e.~12,351 sentences comprising 350,931 tokens. The annotation process consisted in a manual correction and validation of the output of a rule- and heuristics-based named entity recognition and linking tool in an XML editor. 
Only a single person annotated the corpus, despite the limitations of such a protocol, as acknowledged by \newcite{sagot:hal-00703108}.

In total, 5,890 of the 12,351 sentences contain at least a named entity mention. 11,636 mentions were annotated, which are distributed as follows:
3,761 location names, 3,357 company names, 2,381 organization names, 2,025 person names, 67 product names, 29 fiction character names and 15 points of interest.

\subsection{Alignment to the UD version of the FTB}
\label{subsec:alignment}

The named entity (NE) annotation layer for the FTB was developed using an XML editor on the raw text of the FTB. Annotations are provided as inline XML elements within the sentence-segmented but non tokenized text. For creating our NER models, we first had to align these XML annotations with the already tokenized UD version of FTB.

Sentences were provided in the same order for both corpora, so we did not have to align them.
For each sentence, we created a mapping $M$ between the raw text of the NE-annotated FTB (i.e.~after having removed all XML annotations) and tokens in the UD version of the FTB corpus. More precisely, character offsets in the FTB-NE raw text were mapped to token offsets in the tokenized FTB-UD.
This alignment was done using case insensitive character-based comparison and were a mapping of a span in the raw text to a span in the tokenized corpus.
We used the inlined XML annotations to create offline, character-level NE annotations for each sentence, and reported the NE annotations at the token level in the FTB-UD using the mapping $M$ obtained.

We logged each error (i.e.~an unaligned NE or token) and then manually corrected the corpora, as those cases were always errors in either corpora and not alignment errors. We found 70 errors in FTB-NE and 3 errors in FTB-UD. Errors in FTB-NE were mainly XML entity problems (unhandled "\&", for instance) or slightly altered text (for example, a missing comma). Errors in FTB-UD were probably some XML artifacts.

\section{Benchmarking NER Models}

\subsection{Brief state of the art of NER}
\label{subsec:sota}

As mentioned above, NER was first addressed using rule-based approaches, followed by statistical and now neural machine learning techniques. In addition, many systems use a lexicon of named entity mentions, usually called a ``gazetteer'' in this context.

Most of the advances in NER  have been achieved on English, in particular with the CoNLL 2003 \cite{conll03} and  Ontonotes~v5 \cite{pradhan2012conll,pradhan2013towards} corpora. In recent years, NER was traditionally tackled using Conditional Random Fields (CRF) \cite{lafferty2001conditional} which are quite suited for NER; CRFs were later used as decoding layers for Bi-LSTM architectures \cite{huang2015bidirectional,lample2016neural} showing considerable improvements over CRFs alone. These Bi-LSTM-CRF architectures were later enhanced with contextualized word embeddings which yet again brought major improvements to the task \cite{peters2018deep,akbik2018contextual}. Finally, large pre-trained architectures settled the current state of the art showing a small yet important improvement over previous NER-specific architectures \cite{devlin2019bert,baevski2019cloze}.

For French, rule-based system have been developed until relatively recently, due to the lack of proper training data \cite{sekine04,rosset05,stern10np,nouvel11}. The limited availability of a few annotated corpora (cf.~Section~\ref{sec:intro}) made it possible to apply statistical machine learning techniques \cite{bechetcharton_icassp2010,dupont14,dupont2017exploration} as well as hybrid techniques combining handcrafted grammars and machine learning \cite{coop_taln11}. To the best of our knowledge, the best results previously published on FTB NER are those obtained by \newcite{dupont2017exploration}, who trained both CRF and BiLSTM-CRF architectures and improved them using heuristics and pre-trained word embeddings. We use this system as our strong baseline.

Leaving aside French and English, the CoNLL 2002 shared task included NER corpora for Spanish and Dutch corpora \cite{tjong2002introduction} while the CoNLL 2003 shared task included a German corpus \cite{conll03}. The recent efforts by \newcite{strakova2019neural} settled the state of the art for Spanish and Dutch, while \newcite{akbik2018contextual} did so for German.

\begin{table*}
    \ra{1.1}
    \centering\small
    \begin{tabular}{lrrr}
        \toprule
        \textsc{Model} & \textsc{Precision} & \textsc{Recall} & \textsc{F1-Score} \\ 
        \midrule
        \multicolumn{4}{c}{\em baseline}\\
        SEM (CRF) & 87.18 & 80.48 & 83.70\\
        \midrule
        LSTM-seq2seq & 85.10 & 81.87 & 83.45\\
        + FastText & 86.98 & 83.07 & 84.98\\ 
        + FastText + FrELMo & 89.49 & 87.48 & 88.47\\
        + FastText + CamemBERT\textsubscript{OSCAR-BASE-WWM} & 89.79 & 88.86 & 89.32\\
        + FastText + CamemBERT\textsubscript{OSCAR-BASE-WWM} + FrELMo & 90.00 & 88.60 & 89.30\\
        + FastText + CamemBERT\textsubscript{CCNET-BASE-WWM} & 90.31 & 89.29 & 89.80\\
        + FastText + CamemBERT\textsubscript{CCNET-BASE-WWM} + FrELMo & 90.11 & 88.86 & 89.48\\
        + FastText + CamemBERT\textsubscript{OSCAR-BASE-SWM} & 90.09 & 89.46 & 89.77\\
        + FastText + CamemBERT\textsubscript{OSCAR-BASE-SWM} + FrELMo & 90.11 & 88.95 & 89.53\\
        + FastText + CamemBERT\textsubscript{CCNET-BASE-SWM} & 90.31 & 89.38 & 89.84\\
        + FastText + CamemBERT\textsubscript{CCNET-BASE-SWM} + FrELMo & 90.64 & 89.46 & \underline{90.05}\\
        + FastText + CamemBERT\textsubscript{CCNET-500K-WWM} & \underline{90.68} & 89.03 & 89.85\\
        + FastText + CamemBERT\textsubscript{CCNET-500K-WWM} + FrELMo & 90.13 & 88.34 & 89.23\\
        + FastText + CamemBERT\textsubscript{CCNET-LARGE-WWM} & 90.39 & 88.51 & 89.44\\
        + FastText + CamemBERT\textsubscript{CCNET-LARGE-WWM} + FrELMo & 89.72 & 88.17 & 88.94\\
        \midrule
        \multicolumn{4}{c}{\em LSTM-CRF + embeddings}\\
        LSTM-CRF & 85.87 & 81.35 & 83.55\\
        + FastText & 88.53 & 84.63 & 86.53\\
        + FastText + FrELMo & 88.89 & 88.43 & 88.66\\
        + FastText + CamemBERT\textsubscript{OSCAR-BASE-WWM} & 90.47 & 88.51 & 89.48\\
        + FastText + CamemBERT\textsubscript{OSCAR-BASE-WWM} + FrELMo & 89.70 & 88.77 & 89.24\\
        + FastText + CamemBERT\textsubscript{CCNET-BASE-WWM} & 90.24 & 89.46 & 89.85\\
        + FastText + CamemBERT\textsubscript{CCNET-BASE-WWM} + FrELMo & 89.38 & 88.69 & 89.03\\
        + FastText + CamemBERT\textsubscript{OSCAR-BASE-SWM} & \textbf{90.96} & \underline{89.55} & \textbf{90.25}\\
        + FastText + CamemBERT\textsubscript{OSCAR-BASE-SWM} + FrELMo & 89.44 & 88.51 & 88.98\\
        + FastText + CamemBERT\textsubscript{CCNET-BASE-SWM} & 90.09 & 88.69 & 89.38\\
        + FastText + CamemBERT\textsubscript{CCNET-BASE-SWM} + FrELMo & 88.18 & 87.65 & 87.92\\
        + FastText + CamemBERT\textsubscript{CCNET-500K-WWM} & 89.46 & 88.69 & 89.07\\
        + FastText + CamemBERT\textsubscript{CCNET-500K-WWM} + FrELMo & 90.11 & 88.86 & 89.48\\
        + FastText + CamemBERT\textsubscript{CCNET-LARGE-WWM} & 89.19 & 88.34 & 88.76\\
        + FastText + CamemBERT\textsubscript{CCNET-LARGE-WWM} + FrELMo & 89.03 & 88.34 & 88.69\\
        \midrule
        \multicolumn{4}{c}{\em fine-tuning}\\
        mBERT & 80.35 & 84.02 & 82.14\\ %
        
        CamemBERT\textsubscript{OSCAR-BASE-WWM}  & 89.36 & 89.18 & 89.27\\
        CamemBERT\textsubscript{CCNET-500K-WWM}  & 89.35 & 88.81 & 89.08 \\
        CamemBERT\textsubscript{CCNET-LARGE-WWM}  & 88.76 & \textbf{89.58} & 89.39\\
        \bottomrule
    \end{tabular}
    \caption{Results on the test set for the best development set scores.}
    \label{tab:results_ordered}
\end{table*}

\begin{table*}
    \ra{1.1}
    \centering\small
    \begin{tabular}{lrrr}
        \toprule
        \textsc{Model} & \textsc{Precision} & \textsc{Recall} & \textsc{F1-Score} \\ 
        \midrule
        \multicolumn{4}{c}{\em shuf 1}\\
        SEM(dev) & 92.96 & 87.84 & 90.33\\
        LSTM-CRF+CamemBERT\textsubscript{OSCAR-BASE-SWM}(dev) & \underline{93.77} & \underline{94.00} & \underline{93.89}\\
        SEM(test) & 91.88 & 87.14 & 89.45\\
        LSTM-CRF+CamemBERT\textsubscript{OSCAR-BASE-SWM}(test) & \textbf{92.59} & \textbf{93.96} & \textbf{93.27}\\
        \midrule
        \multicolumn{4}{c}{\em shuf 2}\\
        SEM(dev) & 91.67 & 85.96 & 88.73\\
        LSTM-CRF+CamemBERT\textsubscript{OSCAR-BASE-SWM}(dev) & \underline{93.15} & \underline{94.21} & \underline{93.68}\\
        SEM(test) & 90.57 & 87.76 & 89.14\\
        LSTM-CRF+CamemBERT\textsubscript{OSCAR-BASE-SWM}(test) & \textbf{92.63} & \textbf{94.31} & \textbf{93.46}\\
        \midrule
        \multicolumn{4}{c}{\em shuf 3}\\
        SEM(dev) & 92.53 & 88.75 & 90.60\\
        LSTM-CRF+CamemBERT\textsubscript{OSCAR-BASE-SWM}(dev) & \underline{94.85} & \underline{95.82} & \underline{95.34}\\
        SEM(test) & 90.68 & 85.00 & 87.74\\
        LSTM-CRF+CamemBERT\textsubscript{OSCAR-BASE-SWM}(test) & \textbf{91.30} & \textbf{92.67} & \textbf{91.98}\\
        \bottomrule
    \end{tabular}
    \caption{Results on the test set for the best development set scores.}
    \label{tab:results_shuffled}
\end{table*}

\subsection{Experiments}

We used SEM \cite{dupont2017exploration} as our strong baseline because, to the best of our knowledge, it was the previous state-of-the-art for named entity recognition on the FTB-NE corpus.
Other French NER systems are available, such as the one given by SpaCy. However, it was trained on another corpus called WikiNER, making the results non-comparable.
We can also cite the system of \cite{stern2012joint}. This system was trained on another newswire (AFP) using the same annotation guidelines, so the results given in this article are not directly comparable. This model was trained on FTB-NE in \newcite{stern2013identification} (table C.7, page 303), but the article is written in French. The model yielded an F1-score of 0.7564, which makes it a weaker baseline than SEM.
We can cite yet another NER system, namely grobid-ner.\footnote{\url{https://github.com/kermitt2/grobid-ner\#corpus-lemonde-ftb-french}} It was trained on the FTB-NE and yields an F1-score of 0.8739. Two things are to be taken into consideration: the tagset was slightly modified and scores were averaged over a 10-fold cross validation. To see why this is important for FTB-NE, see section \ref{subsubsec:shuffling}.

In this section, we will compare our strong baseline with a series of neural models. We will use the two current state-of-the-art neural architectures for NER, namely seq2seq and LSTM-CRFs models. We will use various pre-trained embeddings in said architectures: fastText, \camembert (a French BERT-like model) and FrELMo (a French ELMo model) embeddings.

\subsubsection{SEM}
SEM \cite{dupont2017exploration} is a tool that relies on linear-chain CRFs \cite{lafferty2001conditional} to perform tagging. SEM uses Wapiti \cite{lavergne2010practical} v1.5.0 as linear-chain CRFs implementation. SEM uses the following features for NER:
\begin{itemize}
    \item token, prefix/suffix from 1 to 5 and a Boolean isDigit features in a [-2, 2] window;
    \item previous/next common noun in sentence;
    \item 10 gazetteers (including NE lists and trigger words for NEs) applied with some priority rules in a [-2, 2] window;
    \item a "fill-in-the-gaps" gazetteers feature where tokens not found in any gazetteer are replaced by their POS, as described in \cite{raymond2010reconnaissance}. This features used token unigrams and token bigrams in a [-2, 2] a window.
    \item tag unigrams and bigrams.
\end{itemize}

We trained our own SEM model by using SEM features on gold tokenization and optimized L1 and L2 penalties on the development set. The metric used to estimate convergence of the model is the error on the development set ($1 - accuracy$). Our best result on the development set was obtained using the rprop algorithm, a 0.1 L1 penalty and a 0.1 L2 penalty.

SEM also uses an NE mention broadcasting post-processing (mentions found at least once are used as a gazetteer to tag unlabeled mentions), but we did not observe any improvement using this post-processing on the best hyperparameters on the development set.

\subsubsection{Neural models}

In order to study the relative impact of different word vector representations and different architectures, we trained a number of NER neural models that differ in multiple ways. They use zero to three of the following vector representations: FastText non-contextual embeddings \cite{bojanowski-etal-2017-enriching}, the FrELMo contextual language model obtained by training the ELMo architecture on the OSCAR large-coverage Common-Crawl-based corpus developed by \newcite{ortiz2019asynchronous}, and one of multiple \camembert language models \cite{arxiv19camembert}. \camembert models are transformer-based models based on an architecture similar to that of RoBERTa \cite{liu2019roberta}, an improvement over the widely used and successful BERT model \cite{devlin2019bert}. The \camembert models we use in our experiments differ in multiple ways:
\begin{itemize}
    \item Training corpus: OSCAR (cited above) or CCNet, another Common-Crawl-based corpus \cite{wenzek2019ccnet} classified by language, of an almost identical size ($\sim$32 billion tokens); although extracted using similar pipelines from Common Crawl, they differ slightly in so far that OSCAR better reflects the variety of genre and style found in Common Crawl, whereas CCNet was designed to better match the style of Wikipedia; moreover, OSCAR is freely available, whereas only the scripts necessary to rebuild CCNet can be downloaded freely. For comparison purposes, we also display the results of an experiment using the mBERT multilingual BERT model trained on the Wikpiedias for over 100 languages.
    \item Model size: following \newcite{devlin2019bert}, we use both ``BASE'' and ``LARGE'' models; these models differ by their number of layers (12 vs.~24), hidden dimensions (768 vs.~1024), attention heads (12 vs.~16) and, as a result, their number of parameters (110M vs.~340M).
    \item Masking strategy: the objective function used to train a \camembert model is a masked language model objective. However, BERT-like architectures like \camembert rely on a fixed vocabulary of explicitly predefined size obtained by an algorithm that splits rarer words into subwords, which are part of the vocabulary together with more frequent words. As a result, it is possible to use a whole-word masked language objective (the model is trained to guess missing words, which might be made of more than one subword) or a subword masked language objective (the model is trained to guess missing subwords). Our models use the acronyms WWM and SWM respectively to indicate the type of masking they used.
\end{itemize}

We use these word vector representations in three types of architectures:
\begin{itemize}
    \item Fine-tuning architectures: in this case, we add a dedicated linear layer to the first subword token of each word, and the whole architecture is then fine-tuned to the NER task on the training data.
    \item Embedding architectures: word vectors produced by language models are used as word embeddings. We use such embeddings in two types of LSTM-based architectures: an LSTM fed to a seq2seq layer and an LSTM fed to a CRF layer. In such configurations, the use of several word representations at the same time is possible, using concatenation as a combination operator. For instance, in Table~\ref{tab:results_ordered}, the model FastText + CamemBERT\textsubscript{OSCAR-BASE-WWM} under the header ``{\em LSTM-CRF + embeddings} corresponds to a model using the LSTM-CRF architecture and, as embeddings, the concatenation of FastText embeddings, the output of the \camembert ``BASE'' model trained on OSCAR with a whole-word masking objective, and the output of the FrELMo language model.
\end{itemize}

For our neural models, we optimized hyperparameters using F1-score on development set as our convergence metric.

We train each model three times with three different seeds, select the best seed on the development set, and report the results of this seed on the test set in Table~\ref{tab:results_ordered}.

\subsubsection{Results}

\paragraph{Word Embeddings:} Results obtained by SEM and by our neural models are shown in table \ref{tab:results_ordered}. First important result that should be noted is that LSTM+CRF and LSTM+seq2seq models have similar performances to that of the SEM (CRF) baseline when they are not augmented with any kind of embeddings. Just adding classical fastText word embeddings dramatically increases the performance of the model.

\paragraph{ELMo Embeddings:} Adding contextualized ELMo embeddings increases again the performance for both architectures. However we note that the difference is not as big as in the case of the pair with/without fastText word embeddings for the LSTM-CRF. For the seq2seq model, it is the contrary: adding ELMo gives a good improvement while fastText does not improve the results as much.

\paragraph{\camembert Embeddings:} Adding the \camembert embeddings always increases the performance of the model LSTM based models. However, as opposed to adding ELMo, the difference with/without \camembert is equally considerable for both the LSTM-seq2seq and LSTM-CRF. In fact adding \camembert embeddings increases the original scores far more than ELMo embeddings does, so much so that the state-of-the-art model is the LSTM + CRF + FastText + CamemBERT\textsubscript{OSCAR-BASE-SWM}.

\paragraph{\camembert + FrELMo:} Contrary to the results given in \newcite{strakova2019neural}, adding ELMo to \camembert did not have a positive impact on the performances of the models. Our hypothesis for these results is that, contrary to \newcite{strakova2019neural}, we trained ELMo and \camembert on the same corpus. We think that, in our case, ELMo either does not bring any new information or even interfere with \camembert.

\paragraph{Base vs large:} an interesting observation is that using large model negatively impacts the performances of the models. One possible reason could be that, because the models are larger, the information is more sparsely distributed and that training on the FTB-NE, a relatively small corpus, is harder.

\subsubsection{Impact of shuffling the data}
\label{subsubsec:shuffling}

One important thing about the FTB is that the underlying text is made of articles from the newspaper Le Monde that are chronologically ordered. Moreover, the standard development and test sets are at the end of the corpus, which means that they are made of articles that are more recent than those found in the training set. This means that a lot of entities in the development and test sets may be new and therefore unseen in the training set. To estimate the impact of this distribution, we shuffled the data, created a new training/development/test split of the same lengths than in the standard split, and retrained and reevaluated our models. We repeated this process 3 times to avoid unexpected biases. The raw results of this experiment are given in table \ref{tab:results_shuffled}. We can see that the shuffled splits result in improvements on all metrics, the improvement in F1-score on the test set ranging from 4.04 to 5.75 (or 25\% to 35\% error reduction) for our SEM baseline, and from 1.73 to 3.21 (or 18\% to 30\% error reduction) for our LSTM-CRF architectures, reaching scores comparable to the English state-of-the-art. This highlights a specific difficulty of the FTB-NE corpus where the development and test sets seem to contain non-negligible amounts of unknown entities. This specificity, however, allows to have a quality estimation which is more in line with real use cases, where unknown NEs are frequent. This is especially the case when processing newly produced texts with models trained on FTB-NE, as the text annotated in the FTB is made of articles around 20 years old.

\section{Conclusion}
\label{sec:conclusion}

In this article, we introduce a new, more usable version of the named entity annotation layer of the French TreeBank. We aligned the named entity annotation to reference segmentation, which will allow to better integrate NER into the UD version of the FTB.

We establish a new state-of-the-art for French NER using state-of-the-art neural techniques and recently produced neural language models for French. Our best neural model reaches an F1-score which is 6.55 points higher (a 40\% error reduction) than the strong baseline provided by the SEM system.

We also highlight how the FTB-NE is a good approximation of a real use case. Its chronological partition increases the number of unseen entities allows to have a better estimation of the generalisation capacities of machine learning models than if it were randomised.

Integration of the NER annotations in the UD version of FTB would allow to train more refined model, either by using more information or through multitask learning by learning POS and NER at the same time. We could also use dependency relationships to provide additional information to a NE linking algorithm.

One interesting point to investigate is that using Large embeddings overall has a negative impact on the models performances. It could be because larger models store information relevant to NER more sparingly, making it harder for trained models to capitalize them. We would like to investigate this hypothesis in future research.

\subsection*{Acknowledgments}

This work was partly funded by the French national ANR grant BASNUM (\mbox{ANR-18-CE38-0003}), as well as by the last author's chair in the PRAIRIE institute,\footnote{\url{http://prairie-institute.fr/}} funded by the French national ANR as part of the ``Investissements d’avenir'' programme under the reference \mbox{ANR-19-P3IA-0001}. The authors are grateful to Inria Sophia Antipolis - Méditerranée ``Nef'' \footnote{\url{https://wiki.inria.fr/wikis/ClustersSophia}} computation cluster for providing resources and support.

\section{Bibliographical References}
\label{main:ref}

\bibliographystyle{lrec}
\bibliography{lrec19ner}

\begin{thebibliography}{}

\bibitem[\protect\citename{Abeill{\'e} \bgroup et al.\egroup }2003]{ftbLR}
Abeill{\'e}, Anne and Cl{\'e}ment, Lionel and Toussenel, Fran{\c c}ois.
\newblock (2003).
\newblock {\em French TreeBank (FTB)}.
\newblock Université Paris-Diderot.

\bibitem[\protect\citename{Candito and Crabb{\'e}}2009]{ftbucLR}
Marie Candito and Benoit Crabb{\'e}.
\newblock (2009).
\newblock {\em French TreeBank with Undone Compounds (FTB-UC)}.
\newblock Université Paris-Diderot.

\end{thebibliography}


\begin{thebibliography}{}

\bibitem[\protect\citename{Akbik \bgroup et al.\egroup
  }2018]{akbik2018contextual}
Akbik, A., Blythe, D., and Vollgraf, R.
\newblock (2018).
\newblock Contextual string embeddings for sequence labeling.
\newblock In {\em Proceedings of the 27th International Conference on
  Computational Linguistics, {COLING} 2018, Santa Fe, New Mexico, USA, August
  20-26, 2018}, pages 1638--1649.

\bibitem[\protect\citename{Baevski \bgroup et al.\egroup
  }2019]{baevski2019cloze}
Baevski, A., Edunov, S., Liu, Y., Zettlemoyer, L., and Auli, M.
\newblock (2019).
\newblock Cloze-driven pretraining of self-attention networks.
\newblock {\em CoRR}, abs/1903.07785.

\bibitem[\protect\citename{Bechet and Charton}2010]{bechetcharton_icassp2010}
Bechet, F. and Charton, E.
\newblock (2010).
\newblock {Unsupervised knowledge acquisition for extracting named entities
  from speech}.
\newblock In {\em 2010 IEEE International Conference on Acoustics, Speech and
  Signal Processing}, Dallas, USA.

\bibitem[\protect\citename{Bojanowski \bgroup et al.\egroup
  }2017]{bojanowski-etal-2017-enriching}
Bojanowski, P., Grave, E., Joulin, A., and Mikolov, T.
\newblock (2017).
\newblock Enriching word vectors with subword information.
\newblock {\em Transactions of the Association for Computational Linguistics},
  5:135--146.

\bibitem[\protect\citename{Béchet \bgroup et al.\egroup }2011]{coop_taln11}
Béchet, F., Sagot, B., and Stern, R.
\newblock (2011).
\newblock Coopération de méthodes statistiques et symboliques pour
  l'adaptation non-supervisée d'un système d'étiquetage en entités
  nommées.
\newblock In {\em Actes de la Conférence TALN 2011}, Montpellier, France.

\bibitem[\protect\citename{Devlin \bgroup et al.\egroup }2019]{devlin2019bert}
Devlin, J., Chang, M., Lee, K., and Toutanova, K.
\newblock (2019).
\newblock {BERT:} pre-training of deep bidirectional transformers for language
  understanding.
\newblock In {\em Proceedings of the 2019 Conference of the North American
  Chapter of the Association for Computational Linguistics: Human Language
  Technologies, {NAACL-HLT} 2019, Minneapolis, MN, USA, June 2-7, 2019, Volume
  1 (Long and Short Papers)}, pages 4171--4186.

\bibitem[\protect\citename{Doddington \bgroup et al.\egroup
  }2004]{doddington2004automatic}
Doddington, G., Mitchell, A., Przybocki, M., Ramshaw, L., Strassel, S., and
  Weischedel, R.
\newblock (2004).
\newblock The automatic content extraction (ace) program-tasks, data, and
  evaluation.
\newblock In {\em Proceedings of LREC - Volume 4}, pages 837--840.

\bibitem[\protect\citename{Dupont and Tellier}2014]{dupont14}
Dupont, Y. and Tellier, I.
\newblock (2014).
\newblock Un reconnaisseur d'entit{\'{e}}s nomm{\'{e}}es du fran{\c{c}}ais.
\newblock In {\em Traitement Automatique des Langues Naturelles, {TALN} 2014,
  Marseille, France, 1-4 Juillet 2014, D{\'{e}}monstrations}, pages 40--41.

\bibitem[\protect\citename{Dupont}2017]{dupont2017exploration}
Dupont, Y.
\newblock (2017).
\newblock Exploration de traits pour la reconnaissance d’entit{\'{e}}s
  nomm{\'{e}}es du fran{\c{c}}ais par apprentissage automatique.
\newblock In {\em 24e Conf{'e}rence sur le Traitement Automatique des Langues
  Naturelles (TALN)}, page~42.

\bibitem[\protect\citename{Galliano \bgroup et al.\egroup }2009]{ester_inter09}
Galliano, S., Gravier, G., and Chaubard, L.
\newblock (2009).
\newblock {The Ester 2 Evaluation Campaign for the Rich Transcription of
  {F}rench Radio Broadcasts}.
\newblock In {\em Interspeech 2009}, Brighton, UK.

\bibitem[\protect\citename{Grouin \bgroup et al.\egroup }2011]{grouin2011law}
Grouin, C., Rosset, S., Zweigenbaum, P., Fort, K., Galibert, O., and Quintard,
  L.
\newblock (2011).
\newblock Proposal for an extension of traditional named entities: From
  guidelines to evaluation, an overview.
\newblock In {\em Proceedings of the Fifth Linguistic Annotation Workshop
  (LAW-V)}, pages 92--100, Portland, OR, June. Association for Computational
  Linguistics.

\bibitem[\protect\citename{Huang \bgroup et al.\egroup
  }2015]{huang2015bidirectional}
Huang, Z., Xu, W., and Yu, K.
\newblock (2015).
\newblock Bidirectional {LSTM-CRF} models for sequence tagging.
\newblock {\em CoRR}, abs/1508.01991.

\bibitem[\protect\citename{Lafferty \bgroup et al.\egroup
  }2001]{lafferty2001conditional}
Lafferty, J.~D., McCallum, A., and Pereira, F. C.~N.
\newblock (2001).
\newblock Conditional random fields: Probabilistic models for segmenting and
  labeling sequence data.
\newblock In {\em Proceedings of the Eighteenth International Conference on
  Machine Learning {(ICML} 2001), Williams College, Williamstown, MA, USA, June
  28 - July 1, 2001}, pages 282--289.

\bibitem[\protect\citename{Lample \bgroup et al.\egroup
  }2016]{lample2016neural}
Lample, G., Ballesteros, M., Subramanian, S., Kawakami, K., and Dyer, C.
\newblock (2016).
\newblock Neural architectures for named entity recognition.
\newblock In {\em {NAACL} {HLT} 2016, The 2016 Conference of the North American
  Chapter of the Association for Computational Linguistics: Human Language
  Technologies, San Diego California, USA, June 12-17, 2016}, pages 260--270.

\bibitem[\protect\citename{Lavergne \bgroup et al.\egroup
  }2010]{lavergne2010practical}
Lavergne, T., Capp{\'e}, O., and Yvon, F.
\newblock (2010).
\newblock Practical very large scale {CRFs}.
\newblock In {\em Proceedings of the 48th Annual Meeting of the Association for
  Computational Linguistics}, pages 504--513. Association for Computational
  Linguistics.

\bibitem[\protect\citename{Liu \bgroup et al.\egroup }2019]{liu2019roberta}
Liu, Y., Ott, M., Goyal, N., Du, J., Joshi, M., Chen, D., Levy, O., Lewis, M.,
  Zettlemoyer, L., and Stoyanov, V.
\newblock (2019).
\newblock Roberta: {A} robustly optimized {BERT} pretraining approach.
\newblock {\em CoRR}, abs/1907.11692.

\bibitem[\protect\citename{Marsh and Perzanowski}1998]{marsh1998muc}
Marsh, E. and Perzanowski, D.
\newblock (1998).
\newblock {MUC}-7 evaluation of {IE} technology: Overview of results.
\newblock In {\em Proceedings of the Seventh Message Understanding Conference
  (MUC-7) - Volume 20}.

\bibitem[\protect\citename{{Martin} \bgroup et al.\egroup
  }2019]{arxiv19camembert}
{Martin}, L., {Muller}, B., {Ortiz Su{\'a}rez}, P.~J., {Dupont}, Y., {Romary},
  L., {Villemonte de la Clergerie}, {\'E}., {Seddah}, D., and {Sagot}, B.
\newblock (2019).
\newblock {CamemBERT: a Tasty French Language Model}.
\newblock {\em arXiv e-prints}, page arXiv:1911.03894, Nov.

\bibitem[\protect\citename{Nouvel \bgroup et al.\egroup }2011]{nouvel11}
Nouvel, D., Antoine, J., and Friburger, N.
\newblock (2011).
\newblock Pattern mining for named entity recognition.
\newblock In {\em Human Language Technology Challenges for Computer Science and
  Linguistics - 5th Language and Technology Conference, {LTC} 2011,
  Pozna{\'{n}}, Poland, November 25-27, 2011, Revised Selected Papers}, pages
  226--237.

\bibitem[\protect\citename{Ortiz~Su{\'a}rez \bgroup et al.\egroup
  }2019]{ortiz2019asynchronous}
Ortiz~Su{\'a}rez, P.~J., Sagot, B., and Romary, L.
\newblock (2019).
\newblock Asynchronous pipeline for processing huge corpora on medium to low
  resource infrastructures.
\newblock {\em Challenges in the Management of Large Corpora (CMLC-7) 2019},
  page~9.

\bibitem[\protect\citename{Peters \bgroup et al.\egroup }2018]{peters2018deep}
Peters, M.~E., Neumann, M., Iyyer, M., Gardner, M., Clark, C., Lee, K., and
  Zettlemoyer, L.
\newblock (2018).
\newblock Deep contextualized word representations.
\newblock In {\em Proceedings of the 2018 Conference of the North American
  Chapter of the Association for Computational Linguistics: Human Language
  Technologies, {NAACL-HLT} 2018, New Orleans, Louisiana, USA, June 1-6, 2018,
  Volume 1 (Long Papers)}, pages 2227--2237.

\bibitem[\protect\citename{Pradhan \bgroup et al.\egroup
  }2012]{pradhan2012conll}
Pradhan, S., Moschitti, A., Xue, N., Uryupina, O., and Zhang, Y.
\newblock (2012).
\newblock {CoNLL-2012} shared task: Modeling multilingual unrestricted
  coreference in ontonotes.
\newblock In {\em Joint Conference on Empirical Methods in Natural Language
  Processing and Computational Natural Language Learning - Proceedings of the
  Shared Task: Modeling Multilingual Unrestricted Coreference in OntoNotes,
  EMNLP-CoNLL 2012, July 13, 2012, Jeju Island, Korea}, pages 1--40.

\bibitem[\protect\citename{Pradhan \bgroup et al.\egroup
  }2013]{pradhan2013towards}
Pradhan, S., Moschitti, A., Xue, N., Ng, H.~T., Bj{\"{o}}rkelund, A., Uryupina,
  O., Zhang, Y., and Zhong, Z.
\newblock (2013).
\newblock Towards robust linguistic analysis using ontonotes.
\newblock In {\em Proceedings of the Seventeenth Conference on Computational
  Natural Language Learning, CoNLL 2013, Sofia, Bulgaria, August 8-9, 2013},
  pages 143--152.

\bibitem[\protect\citename{Raymond and Fayolle}2010]{raymond2010reconnaissance}
Raymond, C. and Fayolle, J.
\newblock (2010).
\newblock Reconnaissance robuste d'entit{\'e}s nomm{\'e}es sur de la parole
  transcrite automatiquement.
\newblock In {\em TALN'10}.

\bibitem[\protect\citename{Rosset \bgroup et al.\egroup }2005]{rosset05}
Rosset, S., Illouz, G., and Max, A.
\newblock (2005).
\newblock {I}nteraction et recherche d'information~: le projet {Ritel}.
\newblock {\em Traitement Automatique des Langues}, 46(3):155--179.

\bibitem[\protect\citename{Rosset \bgroup et al.\egroup }2011]{rosset11}
Rosset, S., Grouin, C., and Zweigenbaum, P.
\newblock (2011).
\newblock Entités nommées structurées~: guide d’annotation {Quaero}.
\newblock Notes et Documents 2011-04, LIMSI, Orsay, France.

\bibitem[\protect\citename{Sagot and Stern}2012]{sagot12aleda}
Sagot, B. and Stern, R.
\newblock (2012).
\newblock Aleda, a free large-scale entity database for {F}rench.
\newblock In {\em Proceedings of LREC}.
\newblock To appear.

\bibitem[\protect\citename{Sagot \bgroup et al.\egroup
  }2012]{sagot:hal-00703108}
Sagot, B., Richard, M., and Stern, R.
\newblock (2012).
\newblock {Annotation r{\'e}f{\'e}rentielle du Corpus Arbor{\'e} de Paris 7 en
  entit{\'e}s nomm{\'e}es}.
\newblock In Georges Antoniadis, et~al., editors, {\em {Traitement Automatique
  des Langues Naturelles (TALN)}}, volume 2 - TALN of {\em Actes de la
  conf{\'e}rence conjointe JEP-TALN-RECITAL 2012}, Grenoble, France, June.

\bibitem[\protect\citename{Sekine and Nobata}2004]{sekine04}
Sekine, S. and Nobata, C.
\newblock (2004).
\newblock {Definition, Dictionaries and Tagger for Extended Named Entity
  Hierarchy}.
\newblock In {\em Proceedings of LREC 2004}, Lisbon, Portugal.

\bibitem[\protect\citename{Stern and Sagot}2010]{stern10np}
Stern, R. and Sagot, B.
\newblock (2010).
\newblock Resources for named entity recognition and resolution in news wires.
\newblock In {\em Proceedings of LREC 2010 Workshop on Resources and Evaluation
  for Identity Matching, Entity Resolution and Entity Management}, La Valette,
  Malte.

\bibitem[\protect\citename{Stern \bgroup et al.\egroup }2012]{stern2012joint}
Stern, R., Sagot, B., and B{\'e}chet, F.
\newblock (2012).
\newblock A joint named entity recognition and entity linking system.
\newblock In {\em Proceedings of the Workshop on Innovative Hybrid Approaches
  to the Processing of Textual Data}, pages 52--60. Association for
  Computational Linguistics.

\bibitem[\protect\citename{Stern}2013]{stern2013identification}
Stern, R.
\newblock (2013).
\newblock {\em Identification automatique d'entit{\'e}s pour l'enrichissement
  de contenus textuels}.
\newblock {Ph.D.} thesis, {Universit\'{e} Paris 7 Denis Diderot}.

\bibitem[\protect\citename{Strakov{\'{a}} \bgroup et al.\egroup
  }2019]{strakova2019neural}
Strakov{\'{a}}, J., Straka, M., and Hajic, J.
\newblock (2019).
\newblock Neural architectures for nested {NER} through linearization.
\newblock In {\em Proceedings of the 57th Conference of the Association for
  Computational Linguistics, {ACL} 2019, Florence, Italy, July 28- August 2,
  2019, Volume 1: Long Papers}, pages 5326--5331.

\bibitem[\protect\citename{Tjong Kim~Sang and De~Meulder}2003]{conll03}
Tjong Kim~Sang, E.~F. and De~Meulder, F.
\newblock (2003).
\newblock Introduction to the conll-2003 shared task: Language-independent
  named entity recognition.
\newblock In {\em Proceedings of CoNLL-2003}, pages pp. 142--147, Edmonton,
  Canada.

\bibitem[\protect\citename{Tjong Kim~Sang}2002]{tjong2002introduction}
Tjong Kim~Sang, E.~F.
\newblock (2002).
\newblock Introduction to the {CoNLL-2002} shared task: Language-independent
  named entity recognition.
\newblock In {\em Proceedings of the 6th Conference on Natural Language
  Learning, CoNLL 2002, Held in cooperation with {COLING} 2002, Taipei, Taiwan,
  2002}.

\bibitem[\protect\citename{{Wenzek} \bgroup et al.\egroup
  }2019]{wenzek2019ccnet}
{Wenzek}, G., {Lachaux}, M.-A., {Conneau}, A., {Chaudhary}, V., {Guzm{\'a}n},
  F., {Joulin}, A., and {Grave}, E.
\newblock (2019).
\newblock {CCNet: Extracting High Quality Monolingual Datasets from Web Crawl
  Data}.
\newblock {\em arXiv e-prints}, page arXiv:1911.00359, Nov.

\end{thebibliography}

\section{Language Resource References}
\label{lr:ref}
\bibliographystylelanguageresource{lrec}
\bibliographylanguageresource{lrec19ner}

\end{document}